# Brain: Biological noise-based logic

**Laszlo B. Kish** [1], **Claes G. Granqvist** [2], **Sergey M. Bezrukov** [3] **and Tamas Horvath** [4,5]

[1] Texas A&M University, Department of Electrical Engineering, College Station, TX 77843-3128, USA;
[2] Department of Engineering Sciences, The Ångström Laboratory, Uppsala University, P.O. Box 534, SE-75121 Uppsala, Sweden;
[3] Laboratory of Physical and Structural Biology, Program in Physical Biology, NICHD, National Institutes of Health, Bethesda, MD 20892, USA;
[4] Fraunhofer IAIS, Schloss Birlinghoven, D-53754 Sankt Augustin, Germany;
[5] Department of Computer Science, University of Bonn, Germany

e-mail: Laszlokish@tamu.edu

**Abstract**
Neural spikes in the brain form stochastic sequences, *i.e.*, belong to the class of pulse noises. This stochasticity is a counterintuitive feature because extracting information—such as the commonly supposed neural information of mean spike frequency—requires long times for reasonably low error probability. The mystery could be solved by noise-based logic, wherein randomness has an important function and allows large speed enhancements for special-purpose tasks, and the same mechanism is at work for the brain logic version of this concept.

### 1. Noise-based logic and brain *versus* computer

Noise-based logic (NBL) [1], which has been inspired by the fact that the neural signals in the brain are stochastic, utilizes independent stochastic processes as well as their superposition to carry the logic signal. A basic *brain logic* version [2] of NBL has been motivated by the following observations:

(*i*) The number of neurons in the brain is similar to the number of switching elements (MOS transistors) in a modern flash drive and about 10% of those of a Macbook Air laptop,

(*ii*) the maximum frequency of neural spikes is about 20 million times less than the clock frequency, and (*iii*) neural spike sequences are stochastic, which suggests that their information channel capacity is further limited.

The above facts indicate that the classical suggestions—that *neural information is statistical* and is carried by the *mean frequency of spikes* or *their cross-correlation* (between spike trains)—are likely to be false. Instead

(*a*) single neural spikes potentially carry orders of magnitude more information than a single bit, and

(*b*) the brain uses a number of special-purpose operations that allow it to achieve *reasonably accurate but not perfect* results in a short time and with relatively small "brain-hardware" and time complexity.

The fact that the brain operates in a different way than a computer can be easily demonstrated. Figure 1 provides an illustrative example: the lines contain strings that are identical with the exception of one line where there is a small difference. The brain detects this different line immediately without reading each character in every line. A computer, however, would scan the image character-by-character and then make a comparison accordingly. If we try carry out the analysis in the computer's way—*i.e.*, reading and comparing each element in the tasks described above—we would perform extremely slowly in a large system. Clearly, the brain is using different schemes than computers and employs various special-purpose, noise-based operations, and it must do so because its "brute force" abilities are weaker than those of a laptop.

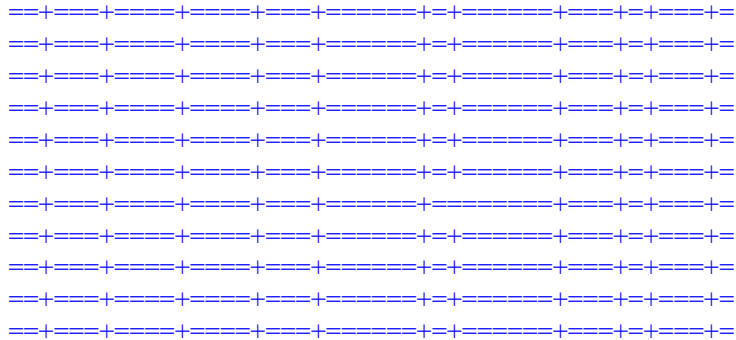

Figure 1: Simple demonstration of the difference between the ways of operation for the brain and a computer.

Two more quick demonstrations of the difference between the brain and a computer are these: We can try to multiply the first 100 integer numbers and check how long does it take; for a laptop computer, it takes less than a millisecond. And we can try to memorize the string "*saenstnsjrmenHixerLöeailarenecltcsrhel*" or its rearranged version "*Hans Liljenström is an excellent researcher*". The second version is much easier for the brain, while it does not matter for the computer. More precisely, the first version is easier for the computer because of the lack of blank characters.

## 2. The essential feature of noise-based brain logic

Due to the limited space, we only illustrate the most essential feature of brain logic [2], which is that the superposition of orthogonal stochastic spike trains carry the information, and the identification of orthogonal components in the superposition is done by coincidence detection (since the neuron is essentially a coincidence detector). As soon as a spike belonging to a component is detected in the superposition by comparing it to the reference signal of the component, its existence is detected, as apparent from Figure 2. Thus, though the signals are stochastic, no time averaging is needed for interpretation of the signal, and the error probability decays exponentially with increasing weighting time.

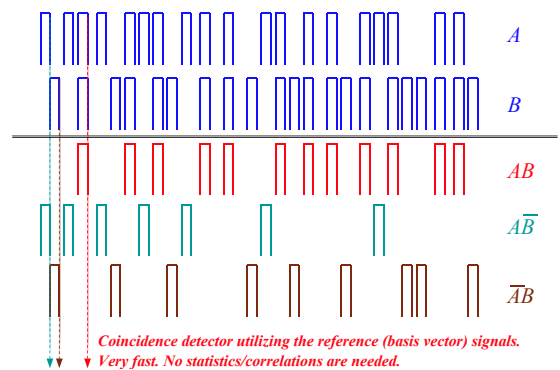

Figure 2. Demonstration why the brain is so fast notwithstanding that neural spikes are stochastic and occur with low frequency. The existence or non-existence of any of the three orthogonal spike trains in superposition sequences *A* and *B* can be quickly observed by coincidence detection, *i.e.*, by neurons.

## 3. Example: String verification by brain logic

Below we show a method on how neurons can solve a string verification problem very rapidly and with non-zero error probability that decreases exponentially *versus* time of operation. The non-brain version of this computational scheme, based on bipolar random telegraph waves, was described in earlier work [3]. In the present paper we propose a brain version and provide the neural circuitry for that, as introduced next.

Suppose that two communicating parts of the brain, called A (Alice) and B (Bob), must verify pairs of *N*-long bit strings via a slow communication channel within the brain. We represent the possible bits in the strings by 2*N* partially overlapping random neural spike sequences (neuro-bits). Via the brain wiring, Alice and Bob have the ability to access these neuro-bits and use them as a reference signal. Then a *hyperspace neural signal* is generated by making the pairwise XOR function of the *N* neuro-bit values of the strings at each clock step. For example, comparing only 83 time steps of the hyperspace signals at Alice and Bob's side provide an error probability of less than $10^{-25}$, *i.e.*, a value of the order of the error rate of regular computer bits [3]. Therefore it is enough to create and communicate a small number of signal bits through the information channel. It is important to note that this error probability is independent of the length *N* of the bit string. Figure 3 shows the neural circuitry to carry out this protocol.

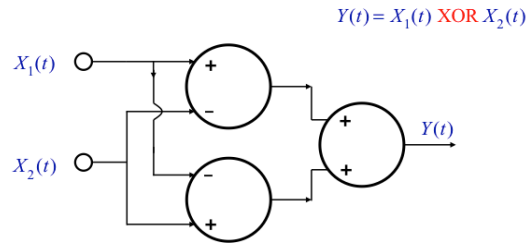

Figure 3. Neural circuitry realizing the XOR logic function for neural spikes. Spherical symbols signify neurons; the "+" and "–" inputs are the excitatory and inhibitory inputs, respectively.

Generalizing this method for the brain may show how intelligence makes reasonable decisions based on a very limited amount of information. Furthermore, our results provide a conceptual explanation why spike transfer via neurons is usually statistical with less than 100% success rate.